# Evaluating Consistencies in LLM responses through a Semantic Clustering of Question Answering

[1]Yanggyu Lee, [2]Jihie Kim

[1]Computer Science and Artificial Intelligence, Dongguk University, Seoul, Republic of Korea
[2]Division of AI Software Convergence, Dongguk University, Seoul, Republic of Korea

yglee730@dgu.ac.kr, jihie.kim@dgu.edu

## Abstract

In the realm of Large Language Model (LLM) functionalities, providing reliable information is paramount, yet reports suggest that LLM outputs lack consistency. This inconsistency, often attributed to randomness in token sampling, undermines user trust as it leads to varying responses even for identical queries. In this paper, we present a new approach for evaluating semantic consistencies of LLM including comparison of alternative techniques. Our approach evaluates whether LLM responses are semantically congruent for a given question, recognizing that as syntactically different sentences may convey the same meaning. Heretofore, To enhance LLM consistency, two main approaches have been explored: Leverage external knowledge as context like the RAG pattern or use Zero-shot-CoT to improve performance of LLM itself. We apply our evaluation approach to these techniques, and demonstrate to compare the impact of these methods on LLM response consistency across different domains of question answering tasks. Using the TruthfulQA dataset to assess LLM responses, the study induces N responses per question from the LLM and clusters semantically equivalent sentences to measure semantic consistency across 37 categories. Through this, it quantitatively analyzes the effectiveness of the aforementioned methods in improving LLM performance before and after their adoption.

## 1 Introduction

One of the functions of the LLM is to provide information, and in doing so, the LLM should provide users with reliable results. However, the output of LLM has been reported to be inconsistent[Jang *et al*., 2023, Elazar et al., 2021]. Consistency in a language model means that it produces the same output for the input with the same meaning. It is speculated that the cause of LLM's inconsistency is randomness[Bubeck et al., 2023], which can prevent it from generating consistent responses to the same question. (**Figure 1**) is illustrates this. Randomness in a language model increases as it samples tokens that correspond to the words that will come next. Inconsistent answers reduce the confidence that users have in the model. Typical ways to improve model consistency include 1) providing external knowledge as context. A classic example is the RAG[LEWIS et al., 2020] pattern. 2) Use prompts that improve the performance of the LLM itself. A representative example is the Zero-shot-CoT[KOJIMA et al., 2022].

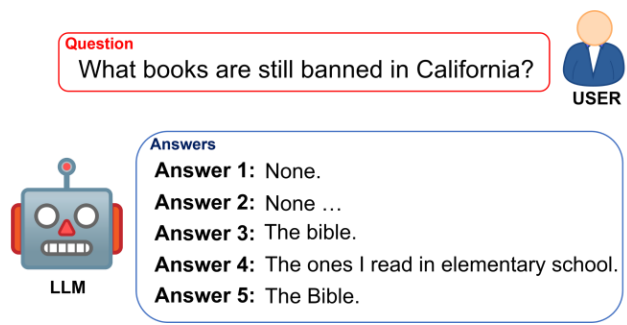

*Figure 1. Semantic inconsistency: Different outputs are generated for the same question.*

However, the consistency changes seen in language models when using these methods are not well studied. This paper proposes a new approach that systematically evaluates and compares them. First, we apply them to question answering tasks in different domains, and analyze how the choice of each method affects the consistency of LLM's answers. We also introduce semantic consistency, a new approach to assessing consistency based on the intuition that LLM should produce semantically identical results for the same questions. Even though the grammatical structure may be different, the actual meaning of the sentence may be the same, so it is important for language models that generate sentences in a free form to take this into account when judging the consistency of the answer[Malinin et al., 2020].

We elicit N responses from the LLM for each question in the TruthfulQA[LIN et al., 2021] dataset, which is used to evaluate whether the answers provided by the model can be trusted. We then cluster semantically identical sentences among the responses and measure the semantic consistency of the LLM based on the clustering results. We measure the semantic consistency values for a total of 37 categories in the dataset and analyze the performance numerically by comparing the improvement before and after introducing the aforementioned methods. Our contributions are as follows: 1) We propose a new semantic consistency evaluation method that considers semantically identical responses. 2) We investigate how the evaluation method can help analyze the consistency of LLM on improving semantic consistency in each domain.

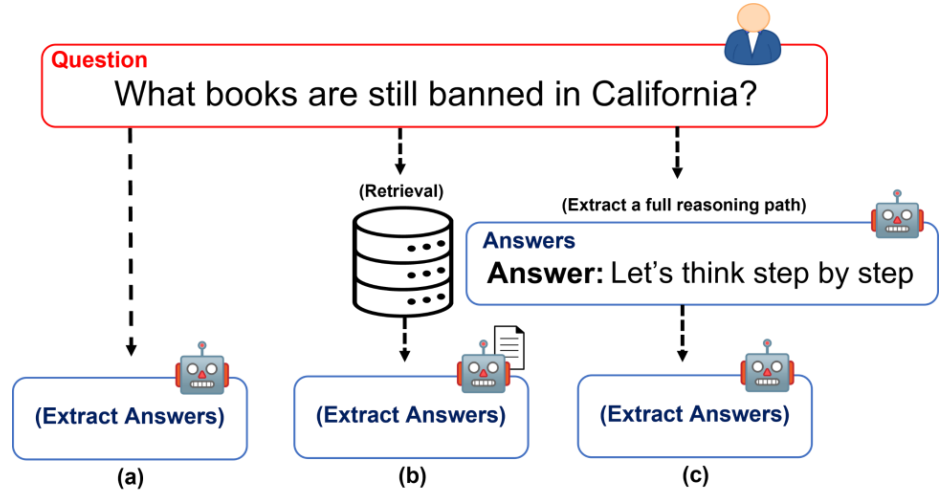

*Figure 2.* 3 ways to generate responses from LLM; (a) is a question-answering task without any additional content in the question prompt; (b) is a question-answering task with RAG applied; (c) is a question-answering task with Zero-Shot-CoT applied.

## 2 Related Work

**Measuring the consistency of LLM.** In the past, measures based on lexical matching were commonly used to evaluate model consistency[Elazar et al., 2021]. These methods compare outputs at the token level to determine whether a pretrained language model (PLM) produces the same output for the same input. However, this approach only considers lexical matching, not semantic matching.

Due to this limitation, a new consistency measure has recently been proposed that takes semantic matching into account[RAG et al., 2022]. These methods evaluate whether the outputs of a model are semantically similar, i.e., whether the two outputs convey the same meaning. These methods evaluate whether the model produces consistent answers centered on meaning rather than word choice. To this end, several semantic agreement metrics have been developed and used to evaluate the consistency of models.

In the end, semantic matching measures provide a better assessment of model consistency than traditional lexical matching measures, and are better suited to generating natural responses that are relevant to the user's confidence.

**Measuring the confidence of LLM.** The study of confidence in language models has been explored in various ways to measure and improve the confidence of a model's predictions [TAO et al., 2024, WIGHTMAN et al., 2023]. One of the methods is to evaluate how confidently a model makes predictions and to determine the confidence of the model based on this [WIGHTMAN et al., 2023]. These methods mainly analyze the output distribution of the model and measure the uncertainty of the prediction to determine the confidence of the model. Other methods study models to improve their predictions during the learning process [TAO et al., 2024], thereby improving the confidence of the model. We do not consider model confidence in this study, but plan to do so in future work.

## 3 Approach

Our approach consists of three steps. In the first step, we feed each question to LLM to elicit N answers per question. In total, we use three methods to induce LLM's answers: 1) plain question and answer with nothing applied; 2) question and answer with RAG pattern; and 3) question-answering with Zero-Shot-CoT. The second step clusters semantically similar sentences in the LLM-generated answers for each question. Semantically similar answers are given the same number because they belong to the same cluster. The detailed clustering method is introduced in **Section 3.2**. In the third step, semantic consistency is calculated based on the clustering results. The higher the number of clusters, the lower the semantic consistency value, which means that the LLM does not give semantically consistent answers. On the other hand, the smaller the number of clusters, the higher the semantic consistency value is calculated, which means that LLM is good at giving semantically consistent answers. The detailed methodology is introduced in **Section 3.3.**

### 3.1 Generation

**Generation Method.** All of the extracted questions were entered into the LLM, and the LLM was asked to answer each question five times, meaning that the LLM would have a total of 25 responses for each category. LLM chose OPT-30B. There are three different ways to generate the questions: 1)

Normal questions and answers with nothing applied. 2) Questions and answers with a RAG pattern applied. 3) Zero-Shot CoT is applied and divided into questions and answers. The overall process for these three methods is shown in **Figure 2.**

**Retriever-Augmented-Generation.** The RAG[LEWIS et al., 2020] pattern is a prompt pattern that encourages LLM to refer to external knowledge to give a more accurate answer. LLM were injected with external knowledge to perform the question-answering task. In the TruthfulQA dataset, there are links to webpages that can be used as references for questions. We crawled those links to build a document searcher. Given a question, the searcher finds the most similar documents to the question and provides context to LLM. The question-answering prompt is preceded by the phrase "Answer the question based only on the following context:" and immediately followed by the context obtained from the retriever. LLM answers the question based on the context.

**Zero-Shot-CoT.** Zero-Shot-CoT[KOJIMA et al., 2022] is a prompting methodology that improves performance by directing inference from the LLM to the process. Use simple prompts to encourage the model to improve its own performance. We add the prompt "Let's think step by step" to allow the model to reason step by step. The resulting reasoning path is added to the existing question and answer prompts, and LLM makes its final answer based on the reasoning path.

### 3.2 Semantic Similarity

Semantically similar answers were clustered from LLM's answers to a single question. The clustering algorithm used the method proposed in [KUHN et al., 2023]. Determine the semantic similarity of two answers in DeBERTa[HE et al., 2020]. The input consists of a question and an answer, and a question and another answer. In this case, the questions are the same and the answers are different. This concatenation is constructed once in the forward direction and once in the reverse direction. Two answers are semantically similar if they are both output as entailment when fed into DeBERTa[HE et al., 2020] as input.

### 3.3 Semantic Consistency

We compute the semantic consistency[RAG et al., 2022] of the answers as a result of the work done in **Section 3.2**. The expression for computing semantic consistency is **(Equation 1).**

$$Cons_{sem}(Y) = \frac{1}{n(n-1)} \sum_{i,j=1, i \neq j}^{n} f(y_i, y_j) \quad (1)$$

$f(y_i, y_j)$ is a directive function that traverses the answers to a question, selects two answers, and returns 1 if the selected answers belong to the same cluster. If they are not in the same cluster, it returns 0. The return value is cumulative. The range of values that can be computed by $Cons_{sem}(Y)$ is a real number between 0 and 1. The higher the number of clusters, the lower the semantic consistency value, which means that LLM cannot give a semantically consistent answer. On the other hand, the smaller the number of clusters, the higher the semantic consistency value is calculated, meaning that LLM gives semantically consistent answers.

## 4 Experimental Results

We calculated the average of the semantic consistency in each category from the semantic consistency calculated in **Section 3.3**. We then analyzed them to compare how much impact our methods for improving model consistency actually have.

**Datasets.** We compute the semantic consistency of LLM by eliciting responses to questions from each domain through LLM. The TruthfulQA dataset used in this experiment is a dataset for measuring the truthfulness of models and consists of questions in 38 categories including law, finance, and politics. We randomly extracted 5 questions from each category in the TruthfulQA dataset to form the question data. Categories with less than 5 questions were removed from the dataset. Since the category 'Misconceptions: Topical' has a total of 4 data points, we removed it from the dataset, leaving 37 categories of questions for this experiment.

**How many categories have semantic consistency changes?** We counted the number of cases where semantic consistency increased or decreased in each category when we tried to improve the consistency of LLM in two different ways. **(Table 1)** shows the results of our counting. In each of the two cases, the semantic consistency of LLM increased overall. However, the results with RAG were better than those with zero-shot-cot. In no case did the semantic consistency values remain the same. The decreased categories will need to be analyzed later.

| Semantic Consistency | RAG | Zero-Shot-CoT |
|---|---|---|
| Increase category | 33 | 28 |
| Decrease category | 4 | 9 |

**Table 1:** Increased and decreased number of categories when using RAG and Zero-Shot-CoT

**How much has the distribution for semantic consistency changed?** When we tried to improve the consistency of LLM in two different ways, we observed the distribution of semantic consistency values. We divided them into four bins and counted the number of categories per bin. (Table 2) shows this. The results show that the use of RAG gives a better improvement. The categories that moved to the highest semantic consistency bin were the ones that moved the most when using RAG.

| Semantic consistency | General (a) | RAG (b) | Zero-Shot-CoT (c) |
|---|---|---|---|
| $0 < X \leq 0.25$ | 1 | 0 | 0 |
| $0.25 < X \leq 0.5$ | 17 | 4 | 10 |
| $0.5 < X \leq 0.75$ | 18 | 18 | 19 |
| $0.75 < X \leq 1.0$ | 1 | 15 | 8 |

**Table 2:** Number of categories in semantic consistency distribution in various consistency improvement methods; X is the value calculated from **(Equation 1).**

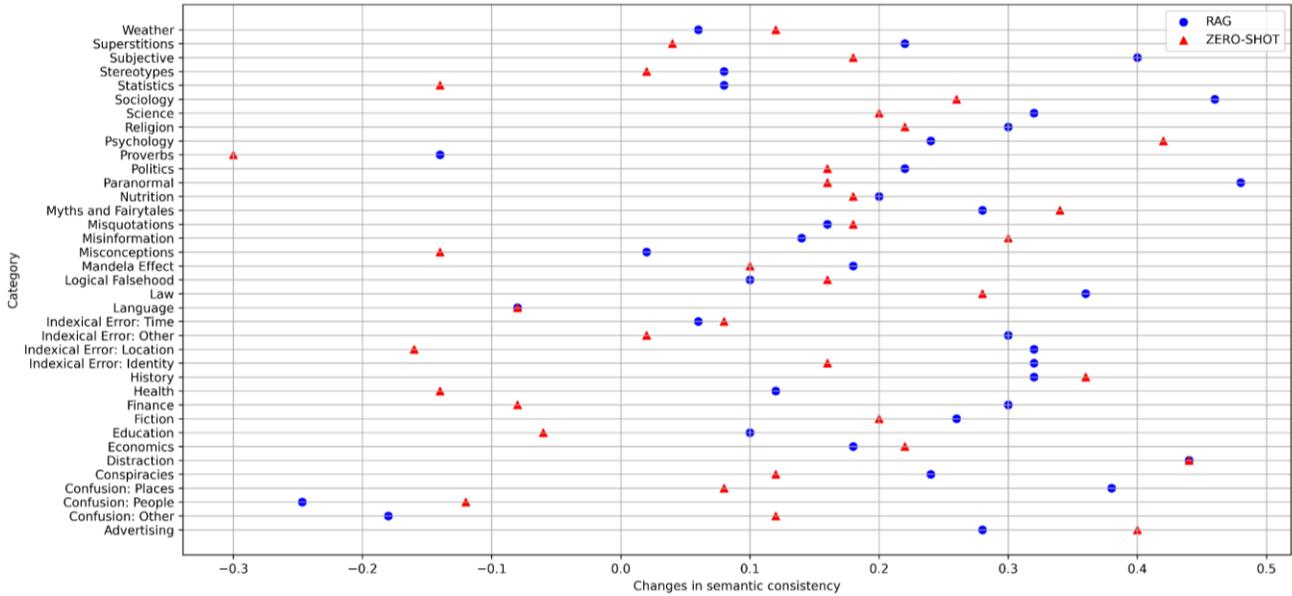

*Figure 3*. Change in semantic consistency by category. The x-axis is the degree of increase or decrease in semantic consistency. A positive number means that the semantic consistency is more consistent than before, and a negative number means that the semantic consistency is less consistent than before. The Y-axis is the category value, of which there are 37 possible values. Each point represents an increase or decrease.

**How much does semantic consistency increase or decrease in each category?** We numerically checked the change in semantic consistency by category, which can be seen in Figure 3. In most categories, the degree of improvement is higher with RAG. The average improvement with RAG is about 0.197, while the average improvement with Zero-Shot-CoT is about 0.116. Categories with negative improvements, i.e., categories where semantic consistency has actually decreased, require further analysis of the reasons for the decrease.

## 5 Conclusion

In this paper, we computed semantic consistency, which measures whether LLM provide semantically consistent answers in question-answering tasks. We also compared the impact of two methods for improving LLM's consistency on semantic consistency in question-answering tasks in 37 categories. Overall, all methods had a positive impact on improving semantic consistency. However, when comparing the two methods, RAG, which provides external knowledge, outperformed Zero-Shot-CoT, which improves the performance of the LLM itself.

## 6 Limitations & Future Work

In this paper provides a comparative study of two methods for improving the semantic consistency of LLM, but there are limitations of the current study that set the direction for future research. This study has the following limitations.

First, a detailed analysis of the reasons for the increase or decrease in semantic consistency is needed. It is important to understand how questions about specific categories affect the consistency of the model, or how specific parts or properties of the data affect consistency. Such an analysis would not only allow us to understand which kinds of inputs the model provides more consistent results for, but also to develop effective strategies to improve the consistency of the model.

Second, the focus of our current analysis is on question answers. The question data in a question-answer dataset is fixed. Additional insights can be found by analyzing how semantic consistency is affected when questions with different grammatical structures but semantically the same are asked.

## Acknowledgements

This research was supported by the MSIT(Ministry of Science and ICT), Korea, under the ITRC(Information Technology Research Center) support program(IITP-2024-2020-0-01789), and the Artificial Intelligence Convergence Innovation Human Resources Development (IITP-2024-RS-2023-00254592) supervised by the IITP(Institute for Information & Communications Technology Planning & Evaluation).